\documentclass[10pt,a4paper,conference]{IEEEtran}

\usepackage{cite}
\usepackage{xcolor}
\usepackage{colortbl}
\usepackage{soul}
\usepackage{graphicx}
\graphicspath{{figures/}}

\begin{document}
\title{Relatable Clothing: Detecting Visual Relationships between People and Clothing}

\author{\IEEEauthorblockN{Thomas Truong}
	\IEEEauthorblockA{\textit{Department of Electrical and Computer Engineering} \\
		\textit{University of Calgary}\\
		\textit{Biometric Technologies Laboratory}\\
		Calgary, Canada \\
		thomas.truong@ucalgary.ca}
	\and
	\IEEEauthorblockN{Svetlana Yanushkevich}
	\IEEEauthorblockA{\textit{Department of Electrical and Computer Engineering} \\
		\textit{University of Calgary}\\
		\textit{Biometric Technologies Laboratory}\\
		Calgary, Canada \\
		syanshk@ucalgary.ca}
}

\maketitle

\begin{abstract}
Detecting visual relationships between people and clothing in an image has been a relatively unexplored problem in the field of computer vision and biometrics. The lack readily available public dataset for ``worn'' and ``unworn'' classification has slowed the development of solutions for this problem. We present the release of the Relatable Clothing Dataset which contains 35287 person-clothing pairs and segmentation masks for the development of ``worn'' and ``unworn'' classification models. Additionally, we propose a novel soft attention unit for performing ``worn'' and ``unworn'' classification using deep neural networks. The proposed soft attention models have an accuracy of upward $98.55\% \pm 0.35\%$ on the Relatable Clothing Dataset and demonstrate high generalizable, allowing us to classify unseen articles of clothing such as high visibility vests as ``worn'' or ``unworn''. 
\end{abstract}

\IEEEpeerreviewmaketitle

\section{Introduction}
The rapid development of object detection and visual relationship detection methodologies has provided the foundation for developing advanced computer vision workplace safety surveillance systems. Object detection involves classifying what classes of objects and localizing their positions in any given image \cite{Felzenszwalb2010}. Visual relationship detection is a lesser researched subject related to object detection. Visual relationship detection involves using object detection results from object detection algorithms and classifying the relationships between all the object pairs in an image \cite{Dai2017}. Many methods, such as YOLOv3 \cite{Redmon2018} and Mask R-CNN \cite{He2017}, have been developed and optimized for object detection and achieve extraordinary results. Conversely, visual relationship detection research is still in its infancy when compared to object detection.

Visual relationship detection for ``worn'' and ``unworn'' clothing suffers from a lack of labeled datasets. In particular, there is currently no publicly available large dataset containing labels pertaining to worn and unworn clothing relationships amongst people-clothing pairs in an image. This is a major detriment as many of the developed solutions for object and visual relationship detection rely on deep learning methods and require large datasets to achieve performance acceptable for industry application. A basic classification system capable of classifying worn and unworn clothing and equipment would provide a strong backbone for advanced workplace safety surveillance systems.

This paper aims to contribute the following to the development and application of state-of-the-art clothing and person relationship detectors for use in industrial settings:
\begin{itemize}
	\item The release of a large dataset, the Relatable Clothing Dataset, labeled for detecting visual relationships between people and worn/unworn clothing.
	\item A performance comparison of several popular feature extraction backbones such as ResNet50 \cite{He2016}, ResNet101 \cite{He2016}, InceptionV3 \cite{Szegedy2016}, and InceptionResNetV2 \cite{Szegedy2016_2} for the visual relationship task of detecting worn and unworn clothing.
	\item A novel model architecture mechanism for soft attention and visual relationship detection when using ResNet backbone networks for feature extraction evaluated quantitatively through the Relatable Clothing Dataset. Additionally, qualitative results are examined for the applicable use case of personal protective equipment usage in the workplace.
\end{itemize}

The paper is organized as follows: Section \ref{sec:literature} covers literature related to visual relationship detection datasets and clothing datasets which could be used for ``worn'' and ``unworn'' people-clothing pair classification. Section \ref{sec:method} details the development of the Relatable Clothing Dataset and our proposed soft attention models. Section \ref{sec:results} summarize the results of our proposed model. Sections \ref{sec:conclusion} and \ref{sec:future} concludes our paper and outlines the  research to be done in the immediate future.

\begin{figure}[h!] % use a * to make figure span across two columns
	\centerline{\includegraphics[width=\linewidth]{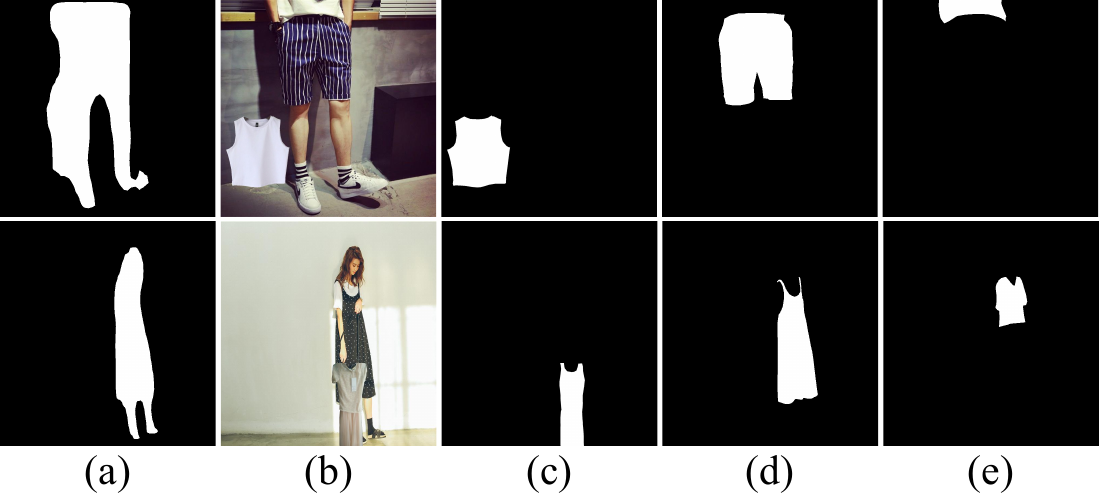}}
	\caption{Example samples from the Relatable Clothing Dataset. From left to right: (a) the person mask, (b) the image, (c) first unworn article of clothing, (d)first worn article of clothing, (e) second worn article of clothing.}
	\label{fig:dataset_examples}
\end{figure} 

\section{Related Works} \label{sec:literature}
\subsection{Visual Relationship Datasets for Clothing}
Verbs in COCO (V-COCO) is the most commonly used dataset for visual relationship detection \cite{gupta2015visual,lin2014microsoft}. This dataset contains labels for people, objects, and verbs such as looking, riding, eating, and drinking. This dataset does not contain any labels for clothing and whether they are worn or unworn. Moreover, the dataset is focused on bounding box labels for objects and relationships (as opposed to instance segmentation labels). Recent works \cite{Kolesnikov2018} and \cite{Gao2019} propose novel attention mechanisms to perform visual relationship detection on this dataset.

Open Images \cite{OpenImages,OpenImagesSegmentation} is a very large dataset containing bounding box labels, instance segmentation labels, and various visual relationship labels. The latest iterations (V5 and V6) of the dataset do contain both bounding boxes and instance segmentation labels for articles of clothing. The dataset also contains the visual relationship of ``wears'' for accessories such as sunglasses, necklaces, purses, and shoes; however, there are no visual relationship labels for larger articles of clothing such as shirts and pants.

DeepFashion2 \cite{Ge2019} is another very large dataset containing bounding box labels and instance segmentation labels specifically for articles of clothing. The dataset contains class labels for top- and bottom-articles of clothing such as short sleeve tops, long sleeve outerwear, shorts, and trousers. Additionally, the dataset provides labels for unworn clothing. Unfortunately, images containing unworn articles of clothing never contain a person in the same image, making the dataset unsuitable for worn/unworn clothing detection in its current state.

Overall, visual relationship detection between objects and people in the field of computer vision has been slow to develop when compared to object detection. A niche application of visual relationship detection can be used in the field of biometrics to detect worn and unworn articles of clothing and protective equipment; however, no dataset exists with the labeled the worn/unworn relationship of various pieces of clothing and protective equipment in an image. \cite{Kolesnikov2018} and \cite{Gao2019} are the most recent works focused on visual relationship classifiers, but neither of these include specifically clothing objects being worn or unworn and operate only on bounding box masks.

The DeepFashion2 dataset can be leveraged to generate images of worn and unworn clothing to train a model. Using results from person and clothing object detectors we can provide a visual relationship classifying network with ``person'' and ``clothing'' object priors which can be used to detect ``worn'' or ``unworn''. Section \ref{sec:dataset_method} outlines the methodology to process the DeepFashion2 dataset to create the Relatable Clothing Dataset.

\section{Methodology} \label{sec:method}
\begin{figure*}[h!] % use a * to make figure span across two columns
	\centerline{\includegraphics[width=\linewidth]{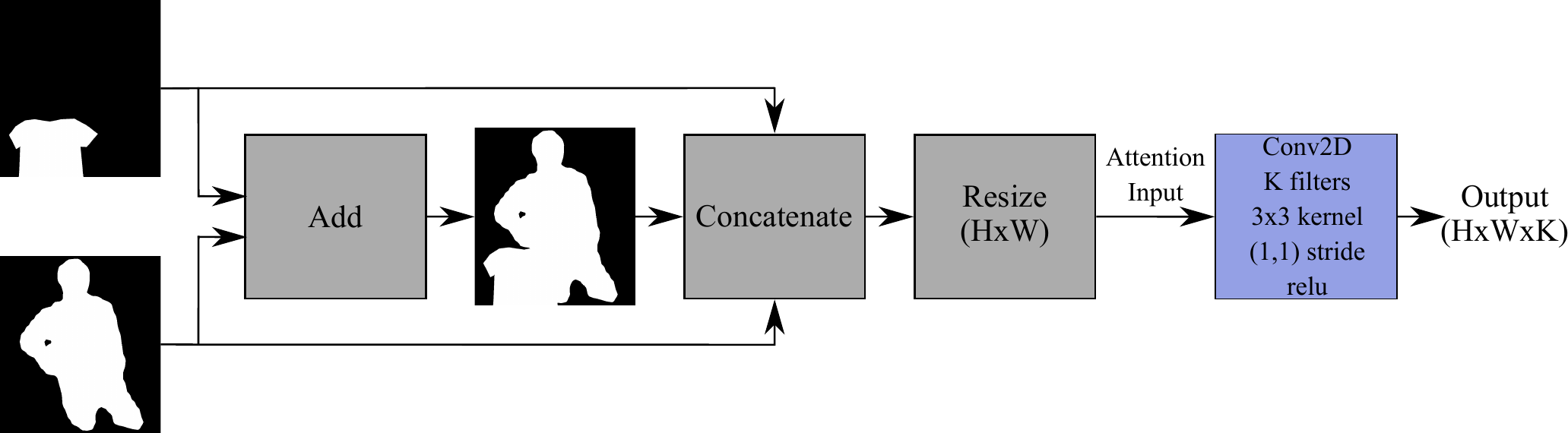}}
	\caption{Proposed soft attention unit. The height (H) and width (W) resize and number of filters (K) matches the output dimensions of the second convolutional layer of each bottleneck unit in the ResNet architecture. Matching output dimensions allows the output of each soft attention unit to be added with the ResNet bottleneck unit convolutional outputs.}
	\label{fig:soft_attention_mask}
\end{figure*} 
We propose to make use of the DeepFashion2 \cite{Ge2019} to create a suitable training and evaluation dataset for classifying people and clothing pairs are worn and unworn. First, we design the Relatable Clothing Dataset from the DeepFashion2 dataset to create samples that contain labeled person and clothing pairs. We then feed our samples from the Relatable Clothing Dataset to a classification model which uses our proposed attention unit to classify relationships between person and clothing pairs as worn or unworn. Finally, we examine the generalizability of our model by conducting a qualitative test case on the practical application of detecting personal protective equipment usage.
\subsection{Relatable Clothing Dataset} \label{sec:dataset_method}

We must first address the lack of availability of a large scale dataset suitable for training and testing a model capable of worn/unworn clothing detection. The Relatable Clothing Dataset is a subset of the DeepFashion2 dataset \cite{Ge2019}. The DeepFashion2 dataset contains segmentation masks for 13 categories of clothing. Additionally, the DeepFashion2 dataset also contains labels for worn/unworn clothing; however, there are a couple caveats preventing direct usage of this dataset to detect ``worn'' and ``unworn'' clothing in person-clothing pairs.

The first caveat is that the dataset contains no segmentation masks for people. This is necessary to create person-clothing pair samples for classification. The second caveat is that most unworn samples are a close-up image of the unworn clothing. There are no images that contain both a person and an unworn piece of clothing.

We overcome the first caveat by deploying a Mask R-CNN \cite{He2017, matterport_maskrcnn_2017} to automatically detect and segment people in the DeepFashion2 dataset. The Mask R-CNN network is pre-trained on the COCO dataset \cite{lin2014microsoft} then fine tuned specifically for person detection on the Open Images V5 dataset \cite{OpenImages,OpenImagesSegmentation} to provide these segmentations. Fig. \ref{fig:dataset_examples}(a) shows example person segmentation masks generated by the network.

The second caveat is overcome by using the segmentation masks for the unworn clothing to crop out their background. The cropped articles of clothing are then overlaid on another image from the dataset, to be called the ``underlay'' image. The ``overlay'' image contains the cropped unworn articles of clothing and is placed in a random (sampled from a uniform random distribution) offset on the ``underlay'' image. Boundary conditions on the uniform random distribution are made such that the ``overlay'' image's original area is atleast 55\% overlapping the ``underlay'' image. The ``underlay'' image contains a person and other articles of clothing (which are worn). Potential overlaps between the unworn article of clothing and the ``underlay'' image clothing may exist. This methodology creates potentially overlapping masks which may contain both unworn and worn clothing sharing the same pixels. These potential overlaps complicate the problem to be more realistic. For example, this can be a rough emulation of a person holding up an unworn article of clothing in such a way that it obscures a portion of their worn articles of clothing.

Fig. \ref{fig:dataset_examples} shows the masks and image examples from the Relatable Clothing Dataset. Each person mask and article of clothing mask pairing is considered to be one sample in the Relatable Clothing Dataset. Formally, the person-clothing mask pair along with its label is considered to be a visual relationship triplet, as defined by the Open Images Dataset \cite{OpenImages}.

Altogether there are 29852 person-clothing pairs (18726 ``worn'' and 11126 ``unworn'') available for training and 5705 person-clothing pairs (3604 ``worn'' and 2101 ``unworn'') for validation and testing. \hl{The dataset and additional technical details can be downloaded here (see supplementary file).} The validation set is evenly partitioned into 10 folds for performance evaluation.

\subsection{Worn and Unworn Classification} \label{sec:worn_method}
We propose a scalable lightweight mask feature extraction network as a soft attention mechanism to be used by the main backbone network architecture. Inspired by the BAR-CNN (Box Attention Relational CNN) paper proposed by Kolesnikov et al. \cite{Kolesnikov2018}, we start with the challenge of detecting the visual relationship triplet, $\langle subject(S),predicate(P), object(O) \rangle $, represented by the probability $p(S,P,O|I)$ where $I$ is a provided image. $P$ describes the relationship between $S$ and $O$. \cite{Kolesnikov2018} simplifies the probability by the chain rule to achieve (\ref{eq:BAR_probability}).

\begin{equation}  \label{eq:BAR_probability}
p(S,P,O|I) = p(S|I) \times p(P,O|S,I)
\end{equation}

We further simplify the problem to focus strictly on predicate classification. We assume independence of $S$ and $O$ (ie., the detection of a subject in an image is independent of the detection of an object in an image) to achieve (\ref{eq:RC_probability}).

\begin{equation}  \label{eq:RC_probability}
p(S,P,O|I) = p(S|I) \times p(O|I) \times p(P|S,O,I)
\end{equation}

The probability $p(S,P,O|I)$ can then calculated in two steps. The first step involves detecting a subject and object pair $\langle subject(S),object(O) \rangle $ in any given $I$. The second step uses the detected subject-object pair, $\langle S,O \rangle $, to a relationship (predicate) classification model to determine $p(P|S,O,I)$ and complete the visual relationship triplet of $\langle subject(S),predicate(P), object(O) \rangle $. For example, we can feed an image through an object detection model to approximate $p(S|I) \approx 0.98$ and $p(O|I) \approx 0.99$. The results from this object detection are then fed to the proposed model in this paper to approximate $p(P|S,O,I) \approx 0.96$. We can then approximate $p(S,P,O|I) \approx 0.980 \times 0.990 \times 0.960 \approx 0.931$, indicating that overall there is a 93.1\% confidence that there is a subject $S$ and object $O$ with predicate $P$ in the given image $I$. This confidence is then usable in risk assessment algorithms in future research. This paper focuses development on the predicate prediction, $p(P|S,O,I)$.

\begin{figure}[h!] % use a * to make figure span across two columns
	\centerline{\includegraphics[width=0.5\linewidth]{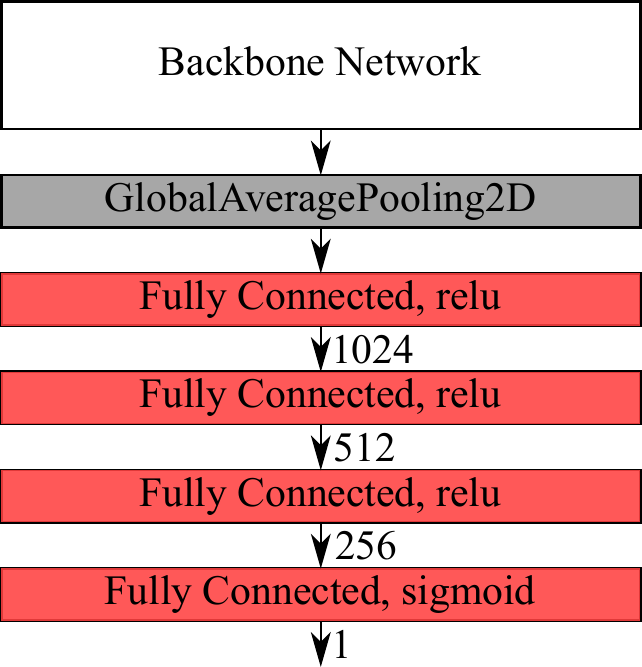}}
	\caption{Network classification head for all trained models in this paper. Each fully connected layer followed by a relu activation contains a 50\% dropout layer to reduce overfitting.}
	\label{fig:network_heads}
\end{figure}

Our proposed model will use the provided ground truth subject-object pairs in the Relatable Clothing Dataset. The dataset contains a binary classification problem as $P = \{worn,unworn\}$. Fig. \ref{fig:soft_attention_mask} is the proposed soft attention architecture. The $\langle S,O \rangle $ masks form the first and second channels of the \textit{Attention Input}. The third channel of the \textit{Attention Input} is the summation of the $\langle S,O \rangle $ masks. The \textit{Attention Input} allows the classification network to determine what $\langle S,O \rangle $ pair is being classified. The \textit{Attention Input} is then resized and passed through a convolutional layer. 

The output of our soft attention unit can then be added with the output of the second convolutional layer of any ResNet bottleneck unit. See \cite{He2016_resnet} for more information on the ResNet architecture and the bottleneck unit. We introduce our soft attention network to the all bottleneck units of both ResNet50 and ResNet101 structures. Our soft attention unit is capable of directing attention towards the  $\langle S,O \rangle $ pair at varying levels of abstraction in the ResNet backbone.

All models in this paper use the classification head depicted in Fig. \ref{fig:network_heads}. The features from the ResNet50 or ResNet101 backbone networks are subsampled with an average pooling layer before being fed to a multilayer perception for classification. 

We quantitatively assess our soft attention unit on our Relatable Clothing dataset. All models in this paper are trained using the training samples from the Relatable Clothing Dataset for 50 epochs and their validation accuracy is monitored using the samples from the first validation partition. The model with the best validation accuracy is kept and then tested on all 10 partitions of the validation samples to provide an idea of model consistency and variance in accuracy, precision, sensitivity, specificity, and F\textsubscript{1} score. Future works using our dataset can directly compare with our results since we provide the Relatable Clothing Dataset with the appropriate partitioning.

A qualitative experiment of the proposed soft attention unit on a sample containing personal protective equipment is also done. The sample is taken from the PPE Dataset provided by \cite{DiBenedettoCBMI2019}. The purpose of this qualitative experiment is to examine if there are any generalizability issues going from the Relatable Clothing Dataset to more practical real world examples with multiple people wearing multiple different articles of clothing, such as personal protective equipment.

An ablation study is used to quantitatively examine the effectiveness of the proposed methodology by removing the proposed mechanisms for improvement. The training and evaluation methodology is maintained and various network architectures and pre-processing are tested.

The first experiment in the ablation study involves reducing the number of soft attention units to 1 (from 16 for the ResNet50 backbone and from 33 for the ResNet101 backbone). This single unit is used with the first bottleneck unit of the ResNet backbones. We want to determine if the reduced complexity from fewer soft attention units significantly impacts model performance.

\begin{figure}[h!] % use a * to make figure span across two columns
	\centerline{\includegraphics[width=0.8\linewidth]{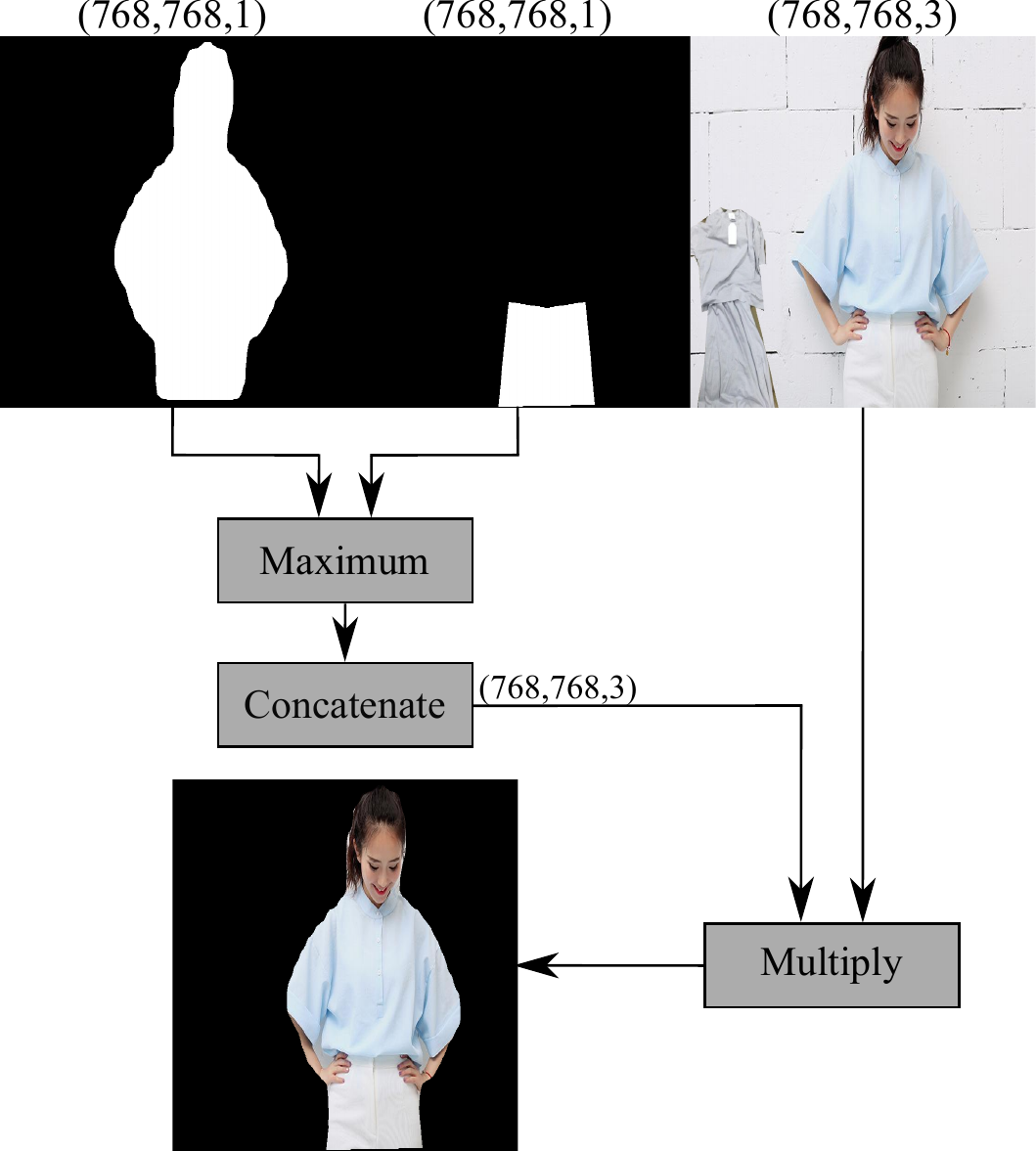}}
	\caption{Input architecture for the baseline hard attention models. The first two inputs contain the masks for the $\langle S,O \rangle$ pair. The third input is the image itself.}
	\label{fig:hard_attention}
\end{figure}

Next, several primitive hard attention models are used as a baseline level of performance for comparison to our soft attention models. The hard attention network is a naive implementation of given prior image masks to create a model capable of determining labels for the $\langle S,O \rangle$ pair priors. Fig. \ref{fig:hard_attention} shows the input network architecture for the hard attention mechanism. The model uses the same network classification head as depicted in Fig. \ref{fig:network_heads}. 

\begin{figure}[h!] % use a * to make figure span across two columns
	\centerline{\includegraphics[width=\linewidth]{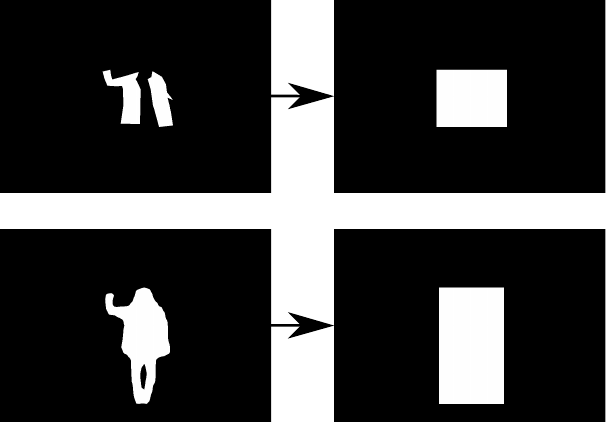}}
	\caption{Segmentation masks are converted to bounding box masks by fitting the smallest possible bounding box to encompass all of the original segmentation masks for both clothing (top) and people (bottom).}
	\label{fig:mask_to_box}
\end{figure} 

Finally, we also examine a soft attention unit which uses bounding box masks instead of segmentation masks for the $\langle S,O \rangle$ pair. This mimics the bounding box masks used by BAR-CNN for V-COCO. Each segmented mask is processed and a bounding box is fit to the segmented masked. The bounding box is sized to contain the smallest possible area while still encompassing the entirety of the segmented mask. Fig. \ref{fig:mask_to_box} shows examples of the mask-to-box conversion. These bounding box masks are then used instead of the segmentation masks for training and testing. 

\section{Results} \label{sec:results}
\begin{table*}[ht!] 
	\centering
	\caption{Performance metrics for the proposed soft attention models.}
	\label{tab:soft_attention_results}
	\begin{tabular}{ccc|ccccc}
		\textbf{\begin{tabular}[c]{@{}c@{}}Soft Attention\\ Backbone\end{tabular}} &
		\textbf{\begin{tabular}[c]{@{}c@{}}Soft Attention\\ Units\end{tabular}} &
		\textbf{\begin{tabular}[c]{@{}c@{}}Trainable\\ Parameters\end{tabular}} &
		\textbf{\begin{tabular}[c]{@{}c@{}}Accuracy\\ (\%)\end{tabular}} &
		\textbf{\begin{tabular}[c]{@{}c@{}}Precision\\ (\%)\end{tabular}} &
		\textbf{\begin{tabular}[c]{@{}c@{}}Recall\\ (\%)\end{tabular}} &
		\textbf{\begin{tabular}[c]{@{}c@{}}Specificity\\ (\%)\end{tabular}} &
		\textbf{\begin{tabular}[c]{@{}c@{}}F\textsubscript{1}\\ (\%)\end{tabular}} \\ \hline
		ResNet50V2 &
		1 &
		26,275,713 &
		$96.00 \pm 1.03$ &
		$98.79 \pm 0.56$ &
		\cellcolor[HTML]{FFFFFF}$94.83 \pm 1.41$ &
		$97.98 \pm 0.96$ &
		\multicolumn{1}{l}{$96.76 \pm 0.85$} \\
		ResNet50V2 &
		16 &
		26,379,649 &
		\cellcolor[HTML]{FFFFFF}$97.74 \pm 0.40$ &
		\cellcolor[HTML]{FFFFFF}$97.76 \pm 0.61$ &
		\cellcolor[HTML]{F8FF00}$98.66 \pm 0.46$ &
		\cellcolor[HTML]{FFFFFF}$96.17 \pm 0.87$ &
		\multicolumn{1}{l}{\cellcolor[HTML]{FFFFFF}$98.21 \pm 0.36$} \\
		ResNet101V2 &
		1 &
		45,285,249 &
		\cellcolor[HTML]{FFFFFF}$97.97 \pm 0.63$ &
		\cellcolor[HTML]{FFFFFF}$98.96 \pm 0.33$ &
		\cellcolor[HTML]{FFFFFF}$97.79 \pm 0.98$ &
		\cellcolor[HTML]{FFFFFF}$98.24 \pm 0.49$ &
		\cellcolor[HTML]{FFFFFF}$98.37 \pm 0.54$ \\
		ResNet101V2 &
		33 &
		45,511,041 &
		\cellcolor[HTML]{F8FF00}$98.55 \pm 0.35$ &
		\cellcolor[HTML]{F8FF00}$99.16 \pm 0.40$ &
		$98.52 \pm 0.50$ &
		\cellcolor[HTML]{F8FF00}$98.58 \pm 0.65$ &
		\cellcolor[HTML]{F8FF00}$98.84 \pm 0.29$
	\end{tabular}
\end{table*}

\begin{table*}[h]
	\centering
	\caption{Performance metrics for the hard attention models.}
	\label{tab:hard_attention_baseline}
	\begin{tabular}{c|ccccl}
		\textbf{Backbone} &
		\textbf{\begin{tabular}[c]{@{}c@{}}Accuracy\\ (\%)\end{tabular}} &
		\textbf{\begin{tabular}[c]{@{}c@{}}Precision\\ (\%)\end{tabular}} &
		\textbf{\begin{tabular}[c]{@{}c@{}}Sensitivity\\ (\%)\end{tabular}} &
		\textbf{\begin{tabular}[c]{@{}c@{}}Specificity\\ (\%)\end{tabular}} &
		\multicolumn{1}{c}{\textbf{\begin{tabular}[c]{@{}c@{}}F\textsubscript{1}\\ (\%)\end{tabular}}} \\ \hline
		ResNet50V2 &
		$92.52 \pm 1.06$ &
		\cellcolor[HTML]{F8FF00}$97.17 \pm 1.00$ &
		$90.79 \pm 1.48$ &
		\cellcolor[HTML]{F8FF00}$95.50 \pm 1.52$ &
		$93.87 \pm 0.91$ \\
		ResNet101V2 &
		\cellcolor[HTML]{F8FF00}$94.11 \pm 0.91$ &
		$95.94 \pm 0.72$ &
		$94.67 \pm 1.25$ &
		$93.17 \pm 0.89$ &
		\cellcolor[HTML]{F8FF00}$95.30 \pm 0.77$ \\
		InceptionV3 &
		$92.59 \pm 0.93$ &
		$94.76 \pm 1.14$ &
		$93.43 \pm 0.96$ &
		$91.17 \pm 1.85$ &
		$94.08 \pm 0.75$ \\
		InceptionResNetV2 &
		$93.51 \pm 0.70$ &
		$94.27 \pm 0.81$ &
		\cellcolor[HTML]{F8FF00}$95.53 \pm 0.82$ &
		$90.04 \pm 1.40$ &
		$94.89 \pm 0.60$
	\end{tabular}
\end{table*}

\begin{table*}[h]
	\centering
	\caption{Performance metrics for the box attention models.}
	\label{tab:box_attention}
	\begin{tabular}{cc|ccccl}
		\textbf{Backbone} &
		\textbf{\begin{tabular}[c]{@{}c@{}}Soft Attention\\ Units\end{tabular}} &
		\textbf{\begin{tabular}[c]{@{}c@{}}Accuracy\\ (\%)\end{tabular}} &
		\textbf{\begin{tabular}[c]{@{}c@{}}Precision\\ (\%)\end{tabular}} &
		\textbf{\begin{tabular}[c]{@{}c@{}}Sensitivity\\ (\%)\end{tabular}} &
		\textbf{\begin{tabular}[c]{@{}c@{}}Specificity\\ (\%)\end{tabular}} &
		\multicolumn{1}{c}{\textbf{\begin{tabular}[c]{@{}c@{}}F\textsubscript{1}\\ (\%)\end{tabular}}} \\ \hline
		\multicolumn{1}{l}{ResNet50V2} &
		1 &
		\multicolumn{1}{l}{$90.99 \pm 1.09$} &
		\multicolumn{1}{l}{$91.90 \pm 1.50$} &
		\multicolumn{1}{l}{$94.04 \pm 0.88$} &
		\multicolumn{1}{l}{$85.78 \pm 2.39$} &
		$92.95 \pm 0.87$ \\
		\multicolumn{1}{l}{ResNet50V2} &
		16 &
		\multicolumn{1}{l}{$93.99 \pm 0.53$} &
		\multicolumn{1}{l}{\cellcolor[HTML]{F8FF00}$98.81 \pm 0.61$} &
		\multicolumn{1}{l}{$91.58 \pm 0.96$} &
		\multicolumn{1}{l}{\cellcolor[HTML]{F8FF00}$98.09 \pm 1.05$} &
		$95.05 \pm 0.52$ \\
		ResNet101V2 &
		1 &
		\cellcolor[HTML]{F8FF00}$95.37 \pm 0.76$ &
		$94.79 \pm 1.04$ &
		\cellcolor[HTML]{F8FF00}$98.04 \pm 0.61$ &
		$90.79 \pm 1.58$ &
		\cellcolor[HTML]{F8FF00}$96.38 \pm 0.66$ \\
		ResNet101V2 &
		33 &
		$95.14 \pm 0.89$ &
		\cellcolor[HTML]{FFFFFF}$97.98 \pm 1.03$ &
		\cellcolor[HTML]{FFFFFF}$94.27 \pm 1.18$ &
		\cellcolor[HTML]{FFFFFF}$96.69 \pm 1.70$ &
		$96.08 \pm 0.71$
	\end{tabular}
\end{table*}

Table \ref{tab:soft_attention_results} summarizes the results for our proposed soft attention models on the Relatable Clothing dataset validation methodology. Sensitivity performance across all models are fairly consistent and indicates that all models perform almost equivalently at detecting worn clothing, with the exception of the ResNet50V2 model with only 1 soft attention unit which performs consierably worse. The addition of more bottleneck units in ResNet101V2 improves precision and specificity over the ResNet50V2 models, indicating that the larger model is less likely to classify unworn clothing as being worn and correctly classifying unworn articles of clothing more often. We also observe that the addition of additional soft attention units mostly sees small improvements across all metrics at the expense of additional training parameters. Highlighted cells indicate the best performance achieved in evaluation.

\begin{figure}[h] % use a * to make figure span across two columns
	\centerline{\includegraphics[width=\linewidth]{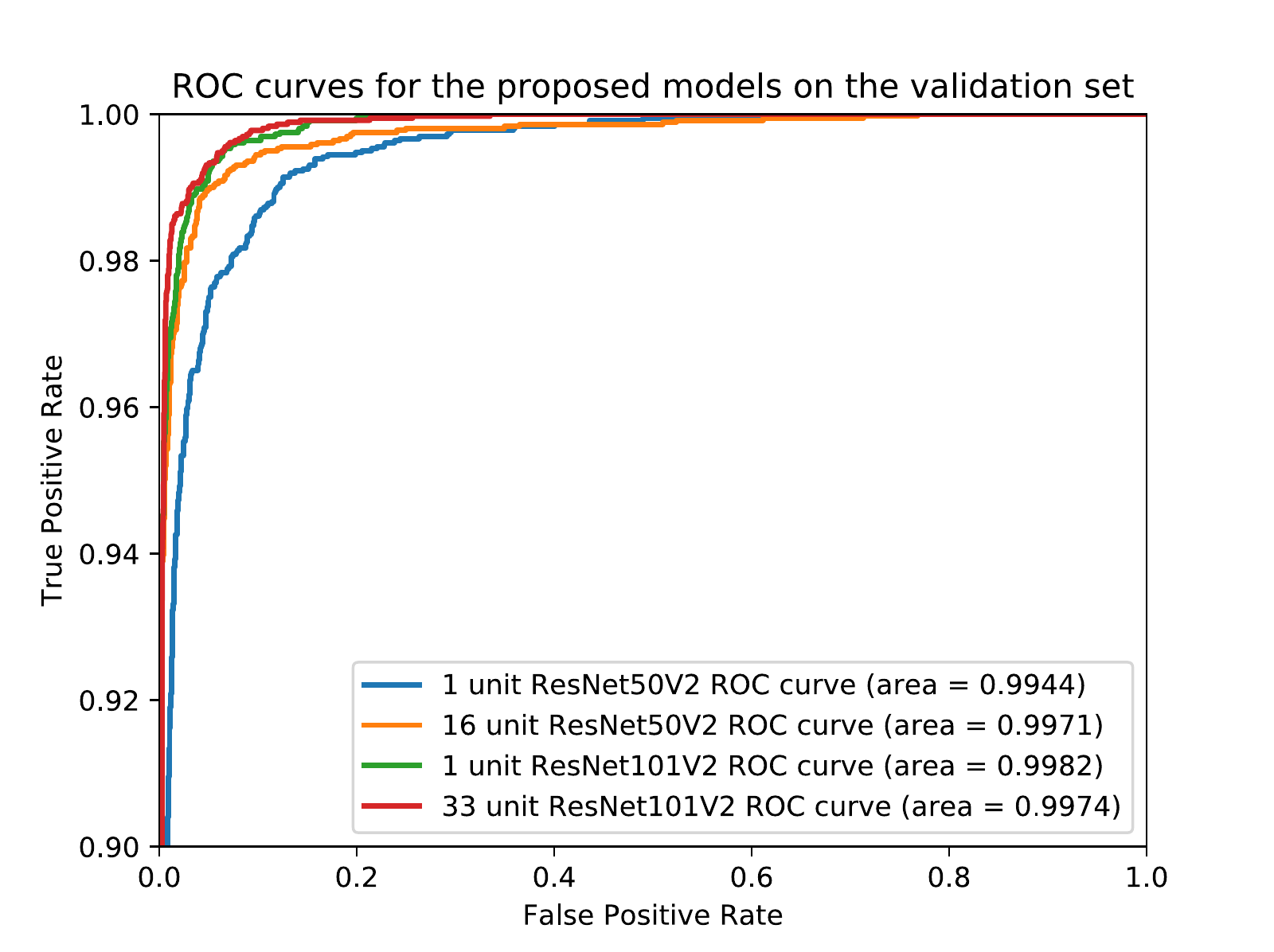}}
	\caption{ROC curves for the proposed soft attention models.}
	\label{fig:ROC_curve}
\end{figure} 

Fig. \ref{fig:ROC_curve} shows the Receiving Operating Characteristics curve for our proposed models on the validation set of the Relatable Clothing Dataset. All models perform very well, with $AUC = 0.9974$ for the 33 soft attention unit ResNet101V2 model. 

\begin{figure*}[h!] % use a * to make figure span across two columns
	\centerline{\includegraphics[width=0.7\linewidth]{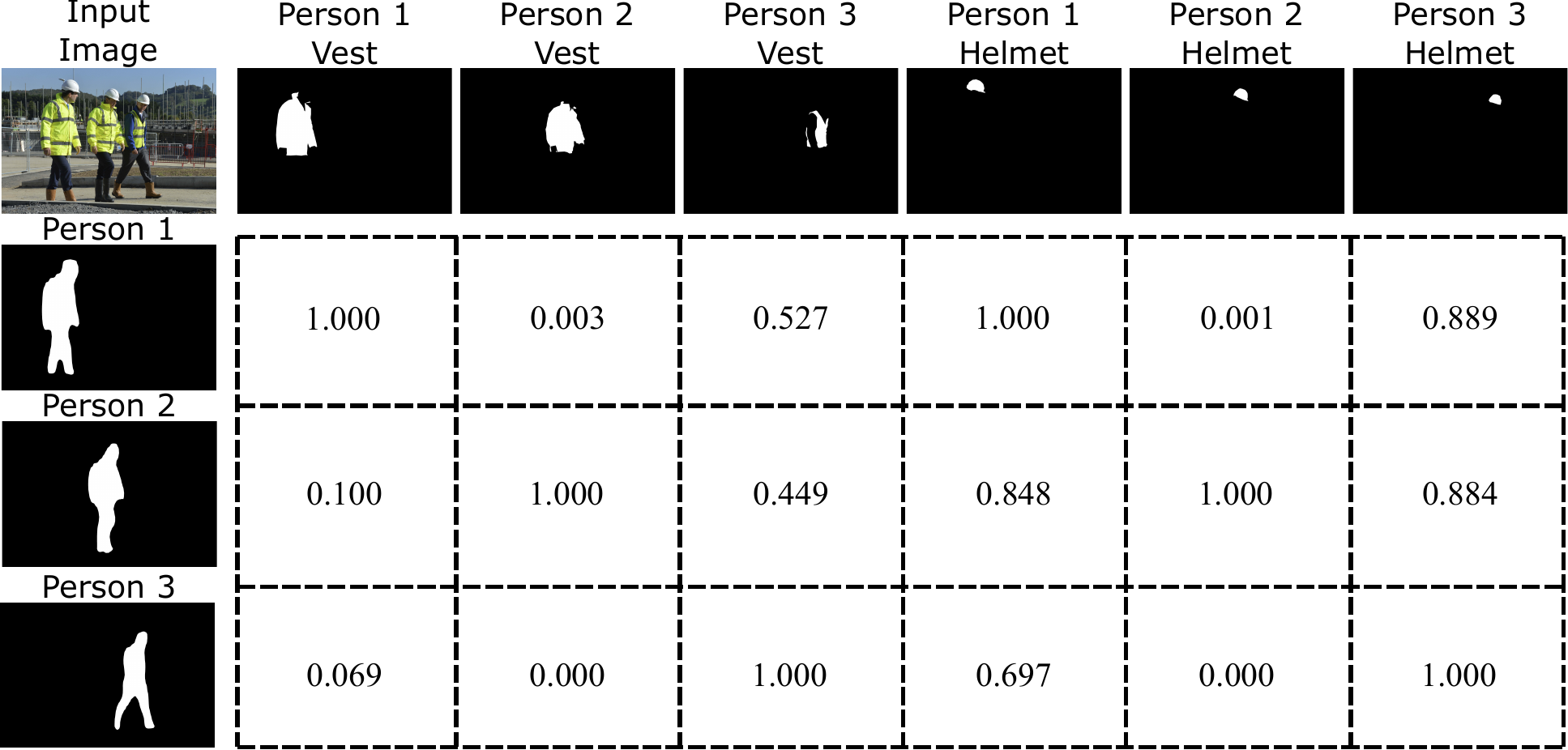}}
	\caption{Qualitative results from our proposed soft attention model on unseen articles of clothing such as high visibility vests and helmets}
	\label{fig:PPE_tests}
\end{figure*}

Fig. \ref{fig:PPE_tests} shows a more qualitative example of the performance of our proposed soft attention models. Along the rows are the person mask inputs and along the columns are the articles of clothing. The value in each cell is the confidence of our proposed model that the article of clothing in the cell column is being worn by the person of the cell row. This confidence can be interpreted as $p(P|S,O,I)$. In total, there are three high visibility vests and three construction helmets being worn in the input image. There are three people, each wearing a single high visibility vest and a single construction helmet. The proposed model has a very high confidence when classifying articles of clothing that are labeled as worn, rounding to $1.000$ at three decimal places and correctly classifying all worn articles of clothing as ``worn''. It is interesting to note that the Relatable Clothing Dataset does not contain any headwear, but all ``worn'' construction helmets are properly classified to their respective person with high confidence. Unfortunately, the model outputs some highly variant confidence values for ``unworn'' articles of clothing and would perform poorly on the ROC curve.

Table \ref{tab:soft_attention_results} shows that increasing the number of soft attention units in the model slightly improves performance at the expense of increased training parameters. Since performance improvements of increased soft attention units are relatively small, we propose using a single soft attention unit for models when training and troubleshooting other model parameters. Additional soft attention units may be added once an optimal model architecture is designed to further improve metrics.

Table \ref{tab:hard_attention_baseline} summarizes the results for the baseline hard attention models. The ResNet backbones perform best in most metrics, but only marginally.

Table \ref{tab:box_attention} summarizes the results for the bounding box soft attention mechanism. These models perform slightly better than the baseline hard attention models but significantly worse than our proposed soft attention models. We also observe similar trends on the box attention models as our proposed soft attention model as we increase the network size.

\section{Conclusion} \label{sec:conclusion}
We proposed a novel soft attention unit for detecting worn and unworn clothing given person and clothing mask priors. In addition, we release a dataset for worn and unworn clothing detection titled Relatable Clothing. We achieve the best performance using a ResNet101V2 backbone with 33 of our proposed soft attention units. This model achieved $98.55\% \pm 0.35\%$ accuracy, $98.58\% \pm 0.65\%$ specificity, and $98.84\% \pm 0.29\%$ F\textsubscript{1} score on the Relatable Clothing Dataset. We also present some promising use cases in workplace safety through qualitative results on a small high-visibility vest dataset. 

\section{Future Works} \label{sec:future}
The works presented in this paper are in the preliminary stages. We look to expand upon our dataset and soft attention unit in the following ways:
\begin{itemize}
	\item Using the proposed soft attention model to the Open Images V6 dataset and creating a semi-automated process for labeling ``worn'' and ``unworn'' person-clothing pairs. The results can then be manually corrected by human annotators to contribute to the Relatable Clothing Dataset.
	\item Expanding our soft attention unit to other visual relationship detection problems from the Open Images V6 dataset and the V-COCO dataset.
	\item Applying the soft attention model to industrial surveillance settings for the detection of proper personal protective equipment usage such as high visibility vests. This would involve potentially expanding the Relatable Clothing Dataset to include additional articles of clothing such as construction helmets, high visibility vests, and ear protection.
\end{itemize}

\section{Acknowledgment}
This project was partially supported by the Natural Sciences and Engineering Research Council of Canada (NSERC) through the grant ``Biometric-enabled Identity Management and Risk Assessment for Smart Cities'' and by the Department of National Defence's Innovation for Defence Excellence and Security (IDEaS) program, Canada.

\bibliographystyle{IEEEtran}
\bibliography{IEEEabrv,references}

\end{document}